%% file: main.tex
\documentclass[10pt,twocolumn,letterpaper]{article}

\usepackage[pagenumbers]{cvpr} 

\input{preamble}

\definecolor{cvprblue}{rgb}{0.21,0.49,0.74}
\usepackage[pagebackref,breaklinks,colorlinks,citecolor=cvprblue]{hyperref}
\usepackage{multirow}
\usepackage{graphicx}
\usepackage{amsmath}
\usepackage{amssymb}
\usepackage{booktabs}
\usepackage{makecell}
\usepackage{footnote}
\usepackage[capitalize]{cleveref}
\crefname{section}{Sec.}{Secs.}
\Crefname{section}{Section}{Sections}
\Crefname{table}{Table}{Tables}
\crefname{table}{Tab.}{Tabs.}

\usepackage{algorithm}
\usepackage{algorithmic}

\usepackage{amsfonts}
\usepackage{latexsym}
\usepackage{amssymb}
\usepackage{amsmath}
\usepackage{amsthm}
\usepackage{booktabs}
\usepackage{enumitem}
\usepackage{graphicx}
\usepackage{color}
\usepackage{overpic}
\usepackage{multirow}
\usepackage{colortbl}
\usepackage{bbding}
\usepackage{pifont}
\usepackage{fontawesome}
\usepackage{subcaption}

\definecolor{lightgrey}{RGB}{244,244,244}
\definecolor{grey}{RGB}{128,128,128}
\definecolor{midgrey}{RGB}{225,225,225}
\definecolor{forestgreen}{RGB}{47, 159, 87}
\newcommand{\cmark}{\color{forestgreen}\ding{51}}%
\newcommand{\xmark}{\color{red}\ding{55}}%

\def\paperID{2627} 
\def\confName{CVPR}
\def\confYear{2025}

\title{RestorerID: Towards Tuning-Free Face Restoration with ID Preservation}

\author{Jiacheng Ying$^{1,\dagger}$ \quad Mushui Liu$^{1,\dagger}$ \quad Zhe Wu$^{1}$ \quad Runming Zhang$^{1}$ \\ Zhu Yu$^{1}$ \quad Siming Fu$^{1}$ \quad Si-Yuan Cao$^{1}$ \quad Chao Wu$^{1}$ \quad Yunlong Yu$^{1}$ \quad Hui-Liang Shen$^{1,*}$\\
{\normalsize $^1$ Zhejiang University}\\
{\tt\small  \{yingjiacheng, lms, jeffw, runmin\_zhang\}@zju.edu.cn} \\
{\tt\small  \{yu\_zhu, fusiming, cao\_siyuan, chao.wu, yuyunlong, shenhl\}@zju.edu.cn}
}

\begin{document}
\input{sec/0_abstract}

\input{sec/1_intro}
\input{sec/2_related_work}

\input{sec/3_method}
\input{sec/4_experiment}

\input{sec/5_conclusion}

{
\small
\bibliographystyle{ieeenat_fullname}
\bibliography{main}
}

\input{sec/6_appendix}


\end{document}

%% file: preamble.tex
\newcommand{\red}[1]{{\color{red}#1}}


%% file: sec/0_abstract.tex
\twocolumn[{
\renewcommand\twocolumn[1][]{#1}
\maketitle
\begin{center}
    \centering
    \vspace*{-.8cm}
    \includegraphics[width=\textwidth]{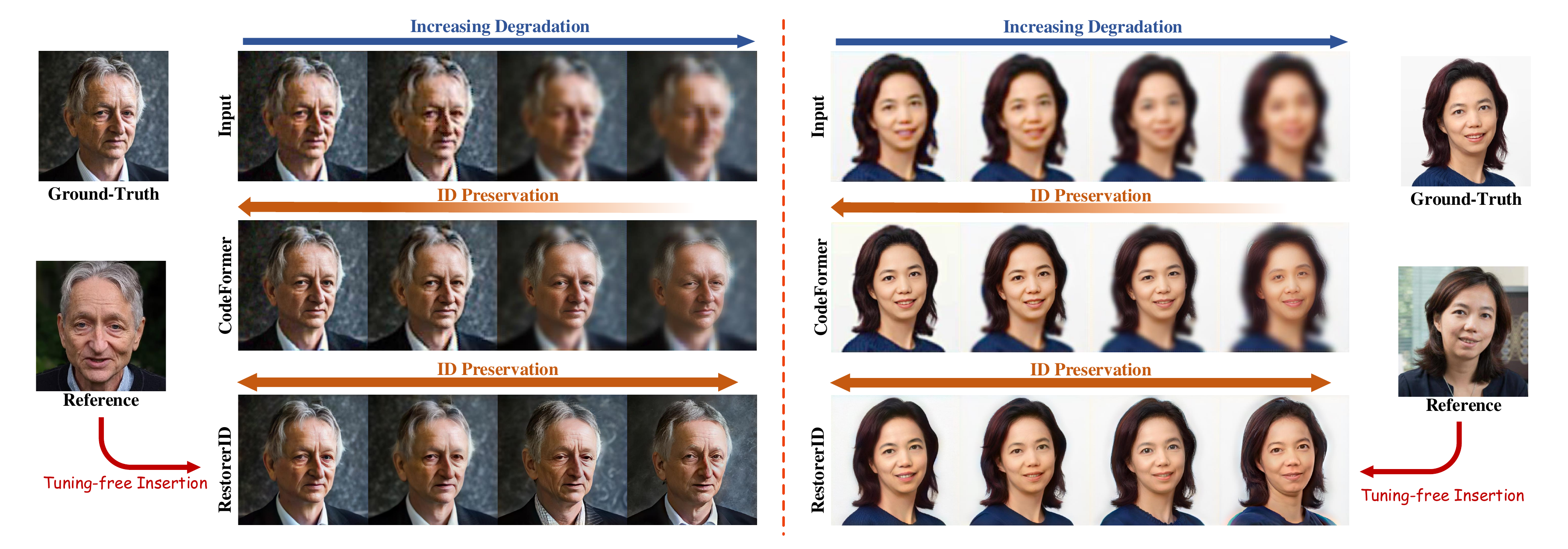}
    \captionof{figure}{As image degradation increases, the blind face restoration approach CodeFormer \cite{CodeFormer2022} can restore images but fails to preserve ID consistency (see the second row). In contrast, Our \textbf{RestorerID}, incorporating reference ID priors, generates restored images with consistent ID information (see the third row).}
    \label{fig:FirstImage}
\end{center}
}]

\let\thefootnote\relax\footnotetext{$\dagger$ Equal contribution. \hspace{3pt} $^*$ Corresponding author.}


\begin{abstract}
Blind face restoration has made great progress in producing high-quality and lifelike images.
Yet it remains challenging to preserve the ID information especially when the degradation is heavy.
Current reference-guided face restoration approaches either require face alignment or personalized test-tuning, which are unfaithful or time-consuming.
In this paper, we propose a tuning-free method named \textbf{RestorerID} that incorporates ID preservation during face restoration.
RestorerID is a diffusion model-based method that restores low-quality images with varying levels of degradation by using a single reference image.
To achieve this, we propose a unified framework to combine the ID injection with the base blind face restoration model.
In addition, we design a novel Face ID Rebalancing Adapter (FIR-Adapter) to tackle the problems of content unconsistency and contours misalignment that are caused by information conflicts between the low-quality input and reference image.
Furthermore, by employing an Adaptive ID-Scale Adjusting strategy, RestorerID can produce superior restored images across various levels of degradation.
Experimental results on the Celeb-Ref dataset and real-world scenarios demonstrate that RestorerID effectively delivers high-quality face restoration with ID preservation, achieving a superior performance compared to the test-tuning approaches and other reference-guided ones. The code of RestorerID is available at \url{https://github.com/YingJiacheng/RestorerID}.
\end{abstract}



%% file: sec/1_intro.tex
\section{Introduction}
\label{sec:intro}
Face restoration \cite{ASFFNet2020, GFRNet2018, DMDnet2022, PFStorer2024} aims to recover clear and high-quality (HQ) facial images from degraded inputs affected by blurring, pixelation, artifacts, JPEG compression, and other noise distortions.
This task has important applications across various fields, \textit{e.g.}, computational photography \cite{SUPIR2024}, old photo recovery \cite{OldPhoto2020}.


To eliminate complex and unknown degradations in low-quality (LQ) images, blind face restoration approaches use GANs \cite{GPEN2021, GFPGAN2021}, codebooks \cite{CodeFormer2022}, and diffusion models \cite{DR22023} to explore general face priors for degradation removal. While these models are capable of generating high-detailed images from LQ inputs, they often struggle to accurately preserve the intricate identity features of human faces. This limitation is particularly evident when dealing with severely degraded images where the ID information are quite unclear, as illustrated in the first and second row in \cref{fig:FirstImage}.

To restore more faithful face images, recent approaches incorporate reference images of the same identity during the restoration process. These approaches can be broadly classified into two categories: the alignment-based one and alignment-free one. Alignment-based approaches, \textit{e.g.}, ASFFNet \cite{ASFFNet2020} and DMDNet \cite{DMDnet2022}, use reference alignment and fusion modules to inject identity features into the restoration process.
However, these approaches often struggle with inaccurate alignment between the reference and LQ input images, leading to suboptimal restoration results. On the other hand, PFStorer \cite{PFStorer2024} introduces an alignment-free approach to learn a neural representation of the identity using personalized models, thereby bypassing the need for direct alignment. While effective in preserving identity without relying on alignment, PFStorer requires model finetuning, taking more than 10 minutes and several (3$\sim$5) images for each identity during testing. This increases computational complexity and limits its practical use. Moreover, this type of test-time fine-tuning often necessitates cloud deployment, which may raise privacy concerns. 



In this work, we explore alignment-free and tuning-free face restoration with ID preservation. This task presents two main challenges: \textbf{\textit{1) How to combine the LQ structural information and reference ID information into a unified framework?}}
Here, the LQ image provides structural information, while the reference image provides ID information. These two types of features should be precisely extracted and uniquely injected into a unified framework.
\textbf{\textit{2) How to reduce the information conflicts between the LQ and reference images?}} Although the reference and LQ images belong to the same identity, they often have significantly differences in illumination, pose, and expression. Direct ID prior injection will disrupt the structure of restored image, and result in a decline of image quality. It is crucial to balance the structural and ID information in the framework.





To address above two challenges, we propose a novel diffusion model-based method, named RestorerID, for ID-preserved face restoration.
For the first challenge, we adopt the independent LQ spatial model and ID model to extract the LQ structural features and ID features, respectively. These two types of features are distinctly injected through ResBlock and Attention modules of a unified diffusion UNet without any parameter conflict. This enables the latent feature to incorporate both the structural and ID information simultaneously during the denoising process. For the second challenge, we propose a Face ID Rebalancing Adapter (FIR-Adapter) that enables an interaction between the LQ structural features and reference ID embeddings, to further enhance the latent features. This adapter effectively avoids facial contours misalignment and content inconsistency during ID injection, significantly improving the image quality while ensuring ID similarity. Additionally, we further design an Adaptive ID-Scale Adjusting strategy based on the level of degradation. This strategy can dynamically adjust the ID injection degree for the model to produce optimal restored images. 
As a result, RestorerID can restore face images with ID preserved across varying levels of degradation, as shown in the third row of \cref{fig:FirstImage}.

\begin{table}[!t]
    \centering
    \resizebox{\linewidth}{!}{
    \begin{tabular}{l|cccc}
    \Xhline{1pt}
    {Method} &{Diffusion-based} & {Ref} & {Tuning-free} & {Alignment-free}\\
    \Xhline{1pt}
    {DR2} \cite{DR22023} & {\cmark} &{\xmark} &{\cmark} &{-}\\
    {CodeFormer} \cite{CodeFormer2022} &{\xmark} & {\xmark} &{\cmark} &{-} \\
    {ASFFNet} \cite{ASFFNet2020} & {\xmark} & {\cmark} &{\cmark} &{\xmark} \\
    {DMDNet} \cite{DMDnet2022} & {\xmark} &{\cmark} &{\cmark} &{\xmark} \\
    {PFStorer} \cite{PFStorer2024} & {\cmark} &{\cmark(3$\sim$5)} &{\xmark} &{\cmark} \\
    \Xhline{1pt}
    {RestorerID}& {\cmark} &{\cmark} &{\cmark} &{\cmark} \\
    \Xhline{1pt}
    \end{tabular}
    }
    \caption{An overview of previous works on face restoration, in comparison with our RestorerID method. }
    \vspace{-3.5mm}
    \label{tab:methods setting}
\end{table}

We distinguish our RestorerID from previous works through its high-fidelity restoration, alignment-free, and tuning-free, along with its superior performance compared to other methods.
\cref{tab:methods setting} concludes the previous face restoration works and provides an overall comparison. Our contributions can be summarized as follows:


\begin{itemize}
    \item We propose a unified framework, RestorerID, a tuning-free method for face restoration that is capable of handling various degradation scenarios while maintaining high-fidelity reconstruction and ID preservation.
    \item We devise the FIR-Adapter to balance the LQ structural information and the reference ID information, improving the restored image quality while maintaining ID preservation. Additionally, we design an Adaptive ID-Scale Adjusting strategy to adaptively generate optimal results according to the level of degradation.
    \item Experimental results validate that RestorerID achieves superior performance compared to the state-of-the-art approaches across different degradation levels.
\end{itemize}

%% file: sec/2_related_work.tex
\section{Related Works} \label{sec:related}

\begin{figure*}[!ht]
    \centering
    \begin{overpic}[width=0.95\linewidth]{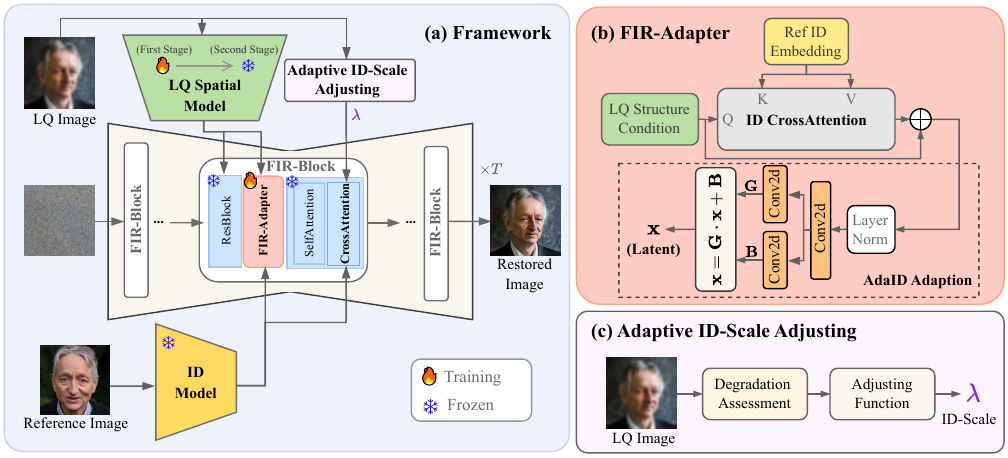}
    \end{overpic}    
    \vspace{-3mm}
    \caption{(a) Our RestorerID framework integrates LQ structural information and reference ID information into a unified diffusion UNet. RestorerID adopts the FIR-Adapter and Adaptive ID-Scale Adjusting module to balance the above two types of information. (b) The FIR-Adapter effectively fuses the LQ structure conditions with reference ID embeddings through an adaptive training mechanism. (c) The Adaptive ID-Scale Adjusting module adjusts the ID injection degree based on degradation assessment.}
    \label{fig:framework}
    \vspace{-3.5mm}
\end{figure*}

\noindent\textbf{Blind Image Restoration.}
Blind face image restoration approaches recover HQ face images from LQ inputs by exploiting face priors, such as geometry, facial textures and colors. Early works use GAN inversion. For example, GPEN \cite{GPEN2021} embeds a GAN within a U-shaped network, followed by fine-tuning for effective blind face restoration. GFP-GAN \cite{GFPGAN2021} employs a GAN framework with carefully designed architectures and losses to leverage generative facial priors, producing high-quality face images. Recent approaches employ diffusion models \cite{DDPM2020} to address severe and unknown degradations in face restoration. DifFace \cite{DifFace2024} creates a transition distribution from LQ images to an intermediate state of a pre-trained diffusion model, gradually recovering HQ images through iterative denoising. DR2 \cite{DR22023} adds Gaussian noise to LQ images, reconstructing HQ targets from the noisy state via a pre-trained diffusion model. PGDiff \cite{PGDiff2023} uses partial guidance to control the denoising process, while BFRffusion \cite{BFRffusion2024} leverages the generative priors in Stable Diffusion, rich in facial and object details, for face restoration. However, under heavy degradation, these approaches struggle to preserve ID information, as critical identity features are often lost during degradation.


\noindent\textbf{ID Preserving Image Generation.}
Despite significant advancements in image generation \cite{stable_diffusion, dalle3}, the field still falls short of meeting the requirements for personalized generation, particularly for human face synthesis. The main challenge lies in the difficulty of enumerating all desired attributes for facial identity generation. This gap has drawn considerable attention to ID-preserving image generation \cite{photomaker,instanceid}, a specialized form of subject-driven generation \cite{anydoor,ssr_encoder,textual_inversion,dreambooth,customdiffusion,ip-adapter}. A subset of methods, such as DreamBooth \cite{dreambooth}, Textual Inversion \cite{textual_inversion}, LoRA \cite{lora}, and ELITE \cite{ELITE}, focus on fine-tuning diffusion models during inference using ID-specific datasets. Meanwhile, recent research has shifted towards training-free ID customization. For instance, IP-Adapter-FaceID \cite{ip-adapter} leverages face ID embeddings from a face recognition model rather than CLIP image embeddings to retain identity consistency. PhotoMaker \cite{photomaker} and Face0 \cite{valevski2023face0} combine text and image embeddings in CLIP space to guide the stable diffusion model, while InstanceID \cite{instantid} explores a plug-and-play module that integrates facial and landmark images with textual prompts to assist in face generation. In this paper, we leverage existing ID-preservation models to aid in face restoration while preserving identity characteristics.

\noindent\textbf{Reference-Guided Face Restoration.}
Reference-guided face restoration aims to improve identity preservation during the restoration process by taking one or several high-quality images of the same identity as guidance. GFRNet \cite{GFRNet2018} explicitly warps the guided face to align with LQ face and further restores corresponding HQ image. ASFFNet \cite{ASFFNet2020} extracts the landmark features as a bridge to fuse the selected guided face image and the LQ image for restoration. DMDNet \cite{DMDnet2022} first memorizes the generic and specific features of facial regions through dual dictionaries, and then adaptively aligns and fuses the relevant features to produce the final result. PFStorer \cite{PFStorer2024} encodes the identity as a neural representation via the test-tuning for personalized face generation, which is quite time-consuming in inference mode. In this paper, we explore tuning-free ID preservation to restore face images even with heavy degradation.

%% file: sec/3_method.tex
\section{Method} \label{sec:method}

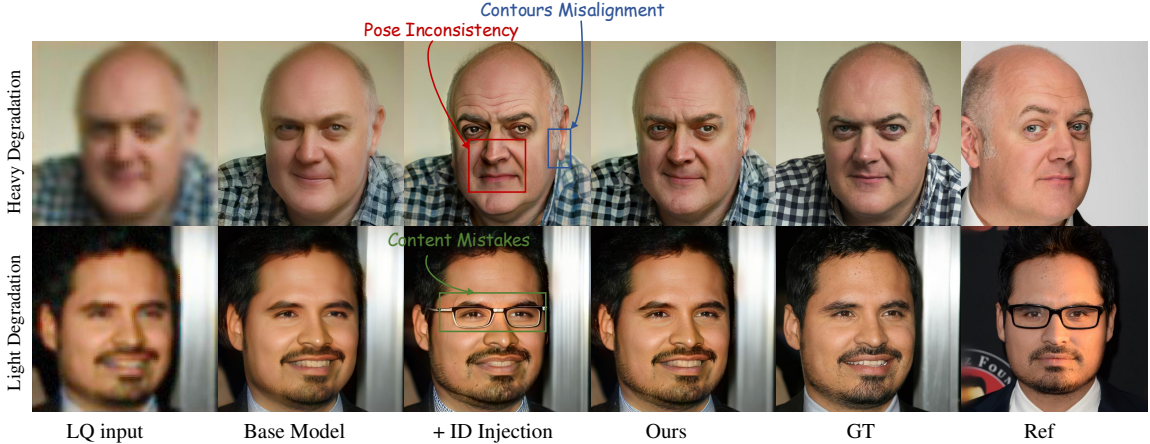
\begin{figure*}[!ht]
    \centering
    \begin{overpic}[width=0.9\linewidth]{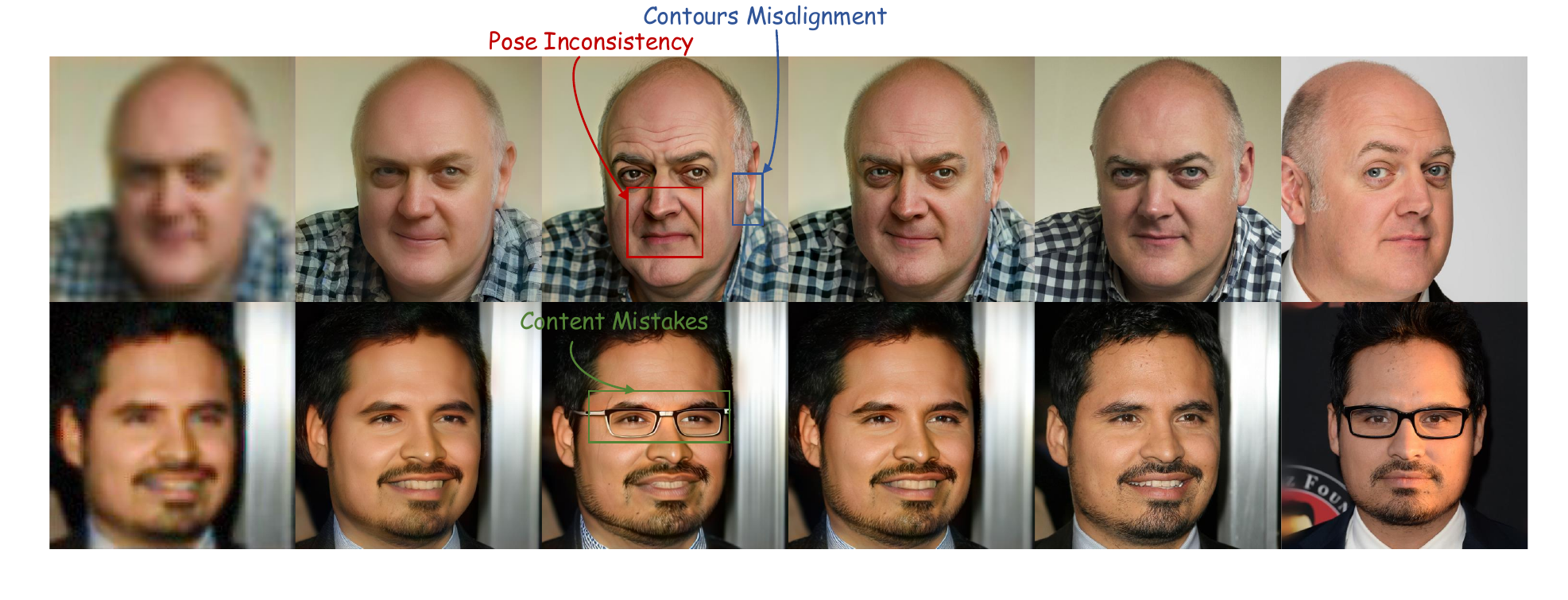}
        \put(6, 1){\footnotesize LQ input}
        \put(21, 1){\footnotesize Base Model}  
        \put(37,1){\footnotesize + ID Injection}
        \put(55, 1){\footnotesize Ours}
        \put(72, 1){\footnotesize GT}
        \put(87, 1){\footnotesize Ref}
        \put(1, 5){\rotatebox{90}{\scriptsize Light Degradation}}
        \put(1, 20){\rotatebox{90}{\scriptsize Heavy Degradation}} 
    \end{overpic}    
    \vspace{-3mm}
    \caption{The outputs produced by the base model, base model + ID injection, and Our method. Blue, red, and green boxes highlight the contours misalignment, pose inconsistency and content mistakes, respectively.}
    \label{fig:ID Injection}
    \vspace{-3.5mm}
\end{figure*}

We propose a tuning-free, ID-preserved face restoration framework, \textbf{RestorerID}, which leverages same-identity reference priors. As shown in \cref{fig:framework} (a), RestorerID consists of key components: the Stable-Diffusion (SD) UNet, LQ Spatial Model, ID Model, FIR-Adapter, and Adaptive ID-Scale Adjusting module. The LQ Spatial Model extracts multi-scale structural features $\mathbf{F}_{\text{lq}}$ to support base restoration, while the ID Model captures reference ID features $\mathbf{F}_{\text{ref}}$, integrated into the UNet through decoupled cross-attention. Positioned between ResBlock and Attention layers, the FIR-Adapter rebalances structural and identity information, with the Adaptive ID-Scale Adjusting module modulating the ID injection degree for optimal results.

\subsection{Preliminaries} \label{sec:preliminaries}
\noindent\textbf{Stable Diffusion (SD) Model.} SD \cite{SD2022} models mainly consist of several components: CLIP \cite{CLIP2021} text encoder for extracting text embeddings, a variational autoencoder (VAE) \cite{VAE2017} with an encoder $\mathcal{E}$ to encode images into a low-dimensional latent space, and a decoder $\mathcal{D}$ to reconstruct images from the latent vectors, and a UNet \cite{UNet2015} for predicting noise during the diffusion process.
The encoder $\mathcal{E}$ maps the input image to the latent space $z_t$, which is efficient and low-dimensional. The optimization is as follows:
\begin{equation}
\mathcal{L}~=~\mathbb{E}_{\varepsilon(\mathbf{x}),\epsilon\sim \mathcal{N}(0,1),t}\parallel\epsilon-\epsilon_{\theta}(\mathbf{z}_{t},t)\parallel_{2}^{2},
 \label{eq:LDM_LOSS}
\end{equation}
where $\epsilon_{\theta}$ is the diffusion model, $\mathbf{z}_{t}$ the latent code at time $t$, and $\epsilon$ the sampled noise.

\noindent\textbf{Image Prompt Adapter.}
Recent ID preservation model, \textit{e.g.}, IP-Adapter \cite{ip-adapter}, adopts the decoupled cross attention to inject the ID embeddings from CLIP image encoder or ID encoder to the SD UNet's Attention module, which can be formulated as:
\begin{equation}
    \mathbf{Z}_{\text{new}} = \text{Attention}(\mathbf{Q}, \mathbf{K}^t, \mathbf{V}^t) + \lambda \cdot \text{Attention}(\mathbf{Q}, \mathbf{K}^i, \mathbf{V}^i), \label{eq:ipa}
\end{equation}
where $\mathbf{Q}$ is mapped from the latent feature, $\mathbf{K}^t$ and $\mathbf{V}^t$ are from the text embeddings, $\mathbf{K}^i$ and $\mathbf{V}^i$ are from the image embeddings, $\lambda$ is the scale weight of the image embeddings.

\subsection{Face Restoration Base Model} \label{sec:face base model}
A strong base model that capable of blind restoration is fundamental to ID-preserved face restoration. Following the setup of PFStorer \cite{PFStorer2024}, we combine the SD with the LQ Spatial Model from StableSR \cite{StableSR2024} as the base model. We retrain the base model using the following optimization:
\begin{equation}
\mathcal{L}~=~\mathbb{E}_{\varepsilon(x),\mathbf{I}_{\text{lq}},\epsilon\sim \mathcal{N}(0,1),t}\parallel\epsilon-\epsilon_{\theta}(\mathbf{z}_{t},\mathbf{I}_{\text{lq}},t)\parallel_{2}^{2},
 \label{eq:Basemodel_LOSS}
\end{equation}
where $\mathbf{I}_{\text{lq}}$ denotes the input LQ image.

\noindent\textbf{Synthetic Degradation.} To obtain HQ-LQ face image pairs, we generate synthetic LQ images by adopting a second-ordered degradation model \cite{RealESRGAN2021} that contains blurring, resizing, noising, and JPEG compression. Furthermore, we improve the noise addition process by converting the image from the sRGB to the raw domain using an ISP model \cite{CBDNet2019}. This operation more closely resembles the real-world noise generation process in camera imaging.

\subsection{ID Preservation} \label{sec:Identity prior injection}
Blind face restoration is an ill-posed problem. When images undergo severe degradation, the ID information, such as facial features, landmarks, and details, are easily compromised. Blind face restoration only relies on general facial priors to restore faces with a similar outline, which is not faithful. In this paper, we incorporate the reference priors from the same identity into the base model to generate more faithful and reliable outputs.

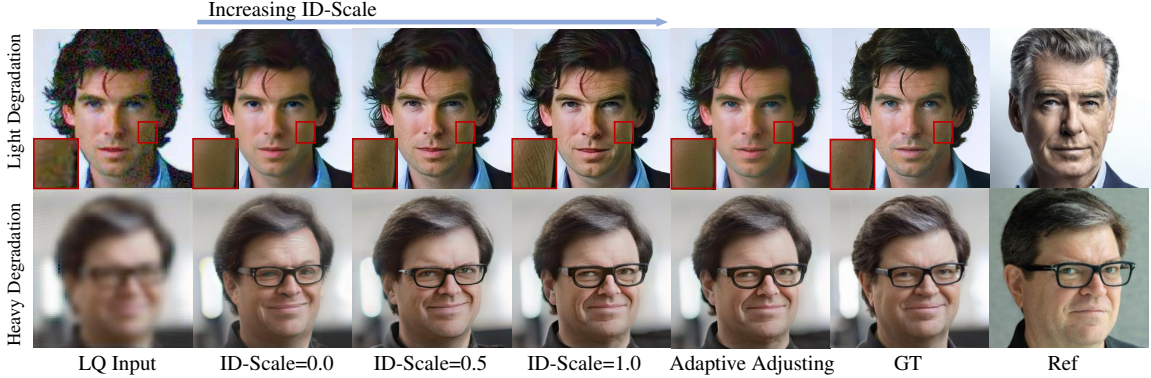
\begin{figure*}[!t]
    \centering
    \begin{overpic}[width=0.9\linewidth]{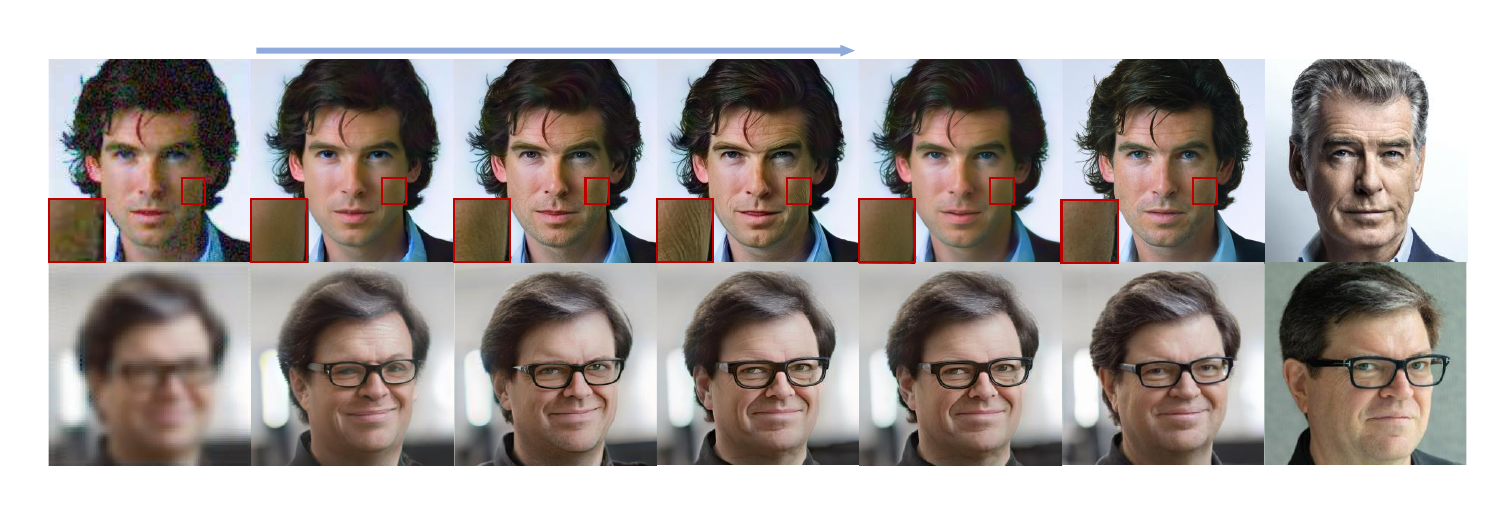}
        \put(7, 1){\footnotesize LQ Input}
        \put(19, 1){\footnotesize ID-Scale=0.0}
        \put(32, 1){\footnotesize ID-Scale=0.5}
        \put(45, 1){\footnotesize ID-Scale=1.0}
        \put(57, 1){\footnotesize Adaptive Adjusting}
        \put(76, 1){\footnotesize GT}
        \put(89, 1){\footnotesize Ref}
        \put(1, 3){\rotatebox{90}{\scriptsize Heavy Degradation}}
        \put(1, 18){\rotatebox{90}{\scriptsize Light Degradation}}        
        \put(18, 31){\footnotesize Increasing ID-Scale}
    \end{overpic}
    \vspace{-2mm}
    \caption{The restored images under light and heavy degradation using increasing ID-Scale values and our adaptive adjusting strategy. Red boxes highlight the facial details. Please zoom in for the best view.}
    \label{fig:IncreasingIPS}
    \vspace{-3.5mm}
\end{figure*}

\noindent\textbf{Impact of ID Injection.} Following IP-Adapter \cite{ip-adapter}, we utilize its pre-trained model (FaceID-Plus) to extract ID embeddings and integrate them through decoupled cross-attention modules, as defined in \cref{eq:ipa}. Further details can be found in the \textit{Supplementary Materials}. However, we observe that direct ID embedding injection can negatively affect restoration results. As shown in \cref{fig:ID Injection}, the first row demonstrates that while the image produced by the base model with ID injection contains more ID information, it struggles to preserve facial contours and pose consistency. In the second row, the image generated by the base model with ID injection exhibits content errors, performing even worse than the base model alone. This is because that the reference and LQ faces have different poses, expressions, and decorations, which means that, the injected embeddings, while incorporating ID priors, also disturb the structural information of the produced images.


\noindent\textbf{FIR-Adapter.} To tackle the aforementioned problem, we design the FIR-Adapter, which is integrated into the UNet to enhance the facial features. As illustrated in \cref{fig:framework} (b), the FIR-Adapter consists of ID CrossAttention and AdaIn Adaption modules. It first enables an interaction between the LQ structural features $\mathbf{F}_{\text{lq}}^{i}$ and reference ID embeddings $\mathbf{F}_{\text{ref}}$ through the cross attention. Then, the FIR-Adapter adopts a LayerNorm and three Convolution layers to obtain the $\mathbf{G}^{i}$ and $\mathbf{B}^{i}$ maps, which are used to enhance the facial details and contour structure of the latent code $\mathbf{x}^{i}$ in a linear manner. The operation of the FIR-Adapter can be formulated as:
\begin{align}
\mathbf{F}^{i}_{\text{en}} &= \mathbf{F}^{i}_{\text{lq}} + \text{Attention}(\mathbf{Q}^{i}_{\text{lq}},\mathbf{K}_{\text{ref}}, \mathbf{V}_{\text{ref}}), \label{eq:Fenhance} \\
\mathbf{G}^{i} &= \text{Conv}(\text{Conv}(\text{LayerNorm}(\mathbf{F}^{i}_{\text{en}}))), \label{eq:Gi} \\
\mathbf{B}^{i} &= \text{Conv}(\text{Conv}(\text{LayerNorm}(\mathbf{F}^{i}_{\text{en}}))), \label{eq:Bi} \\
&\quad \mathbf{x}^{i}_{\text{out}} = \mathbf{G}^{i} \cdot \mathbf{x}^{i} + \mathbf{B}^{i}, \label{eq:Gi_out}
\end{align}
where $\mathbf{Q}^{i}_{\text{lq}}$ is obtained from $\mathbf{F}^{i}_{\text{lq}}$, $\mathbf{K}_{\text{ref}}$ and $\mathbf{V}_{\text{ref}}$ are from $\mathbf{F}_{\text{ref}}$.


\noindent\textbf{Second-stage Training.} We train the FIR-Adapter in the second stage with the other components locked. During the training process, we input LQ and reference images as conditions, and set the ID-Scale $\lambda$ in \cref{eq:ipa} to 0.75. Additionally, we apply random dropout of LQ or reference images to enable classifier-free guidance in the inference stage. We use the modified diffusion loss as follows:
\begin{equation}
\mathcal{L}~=~\mathbb{E}_{\varepsilon(\mathbf{x}),\mathbf{I}_{\text{lq}},\mathbf{I}_{\text{ref}},\epsilon\sim \mathcal{N}(0,1),t}\parallel\epsilon-\epsilon_{\theta}(\mathbf{z}_{t},\mathbf{I}_{\text{lq}},\mathbf{I}_{\text{ref}},t)\parallel_{2}^{2},
 \label{eq:Basemodel_LOSS}
\end{equation}
where $\mathbf{I}_{\text{lq}}$ and $\mathbf{I}_{\text{ref}}$ have probabilities to be zero tensors.

\begin{figure}[!t]
    \centering
    \begin{overpic}[width=1.0\linewidth]{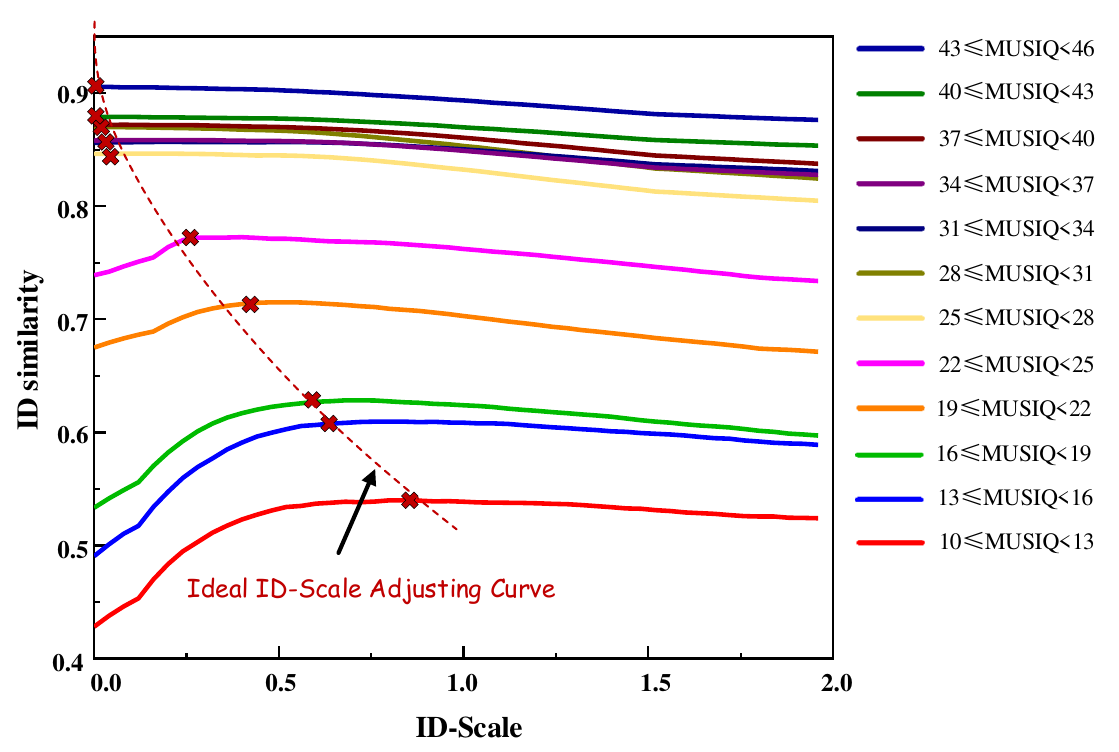}        
    \end{overpic}
    \vspace{-5.5mm}
    \caption{The curves of the average ID similarity values with respect to IP-Scale across different MUSIQ intervals.}
    \vspace{-5.5mm}
    \label{fig:IDScaleCurve}
\end{figure}

\begin{figure*}[!t]
    \centering
    \begin{overpic}[width=0.98\linewidth]{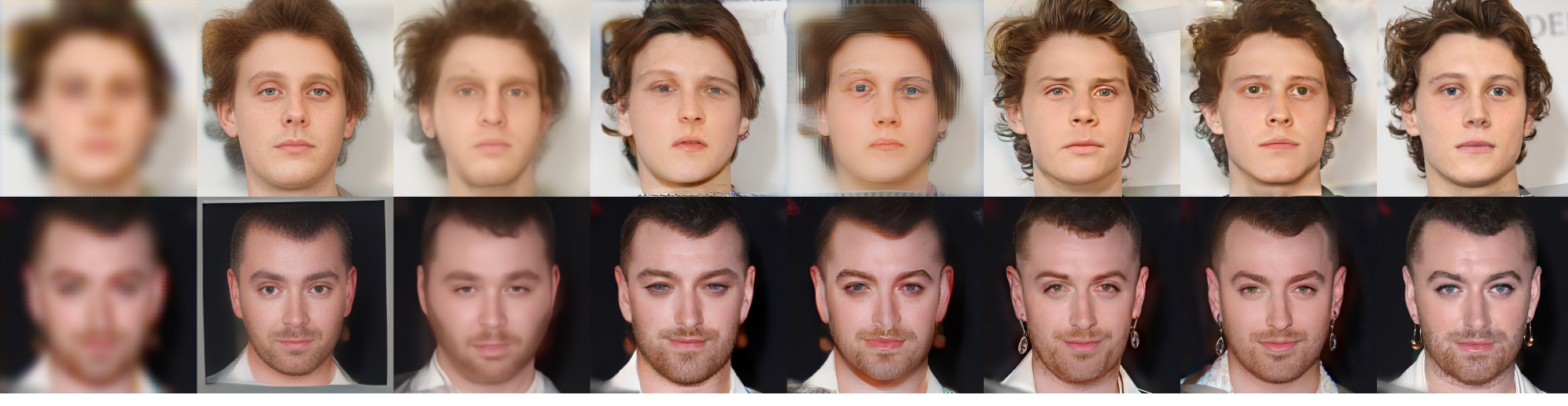}
        \put(4, -1){\footnotesize Input}
        \put(14, -1){\footnotesize DR2+SPAR~\cite{DR22023}}
        \put(26, -1){\footnotesize CodeFormer~\cite{CodeFormer2022}}
        \put(39, -1){\footnotesize ASFFNet~\cite{ASFFNet2020}}
        \put(52, -1){\footnotesize DMDNet~\cite{DMDnet2022}}
        \put(66, -1){\footnotesize PFStorer~\cite{PFStorer2024}}
        \put(80, -1){\footnotesize \textbf{Ours}}
        \put(93, -1){\footnotesize GT}
    \end{overpic}
    \caption{Qualitative comparison with state-of-the-art restoration models on Celeb-Ref dataset with heavy synthetic degradation.}
    \vspace{-1mm}
    \label{fig: CelebRef-Heavy}
\end{figure*}
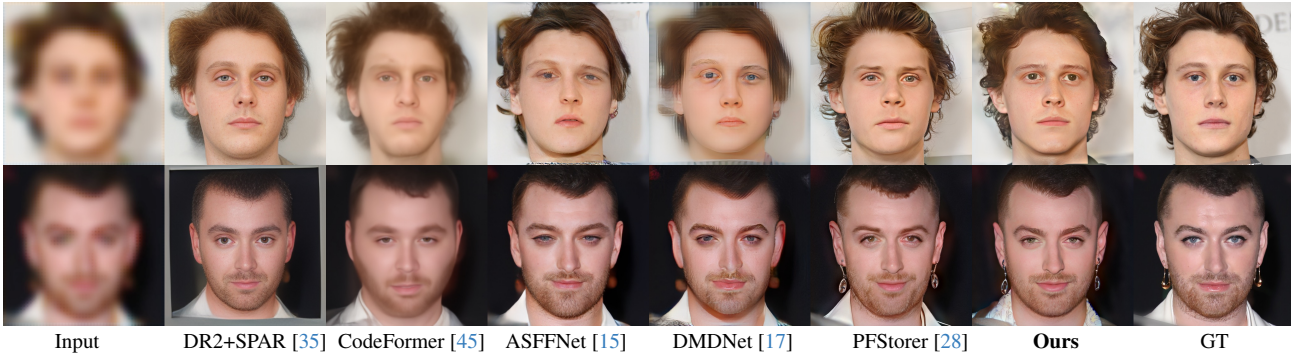

\begin{table*}[!ht]
    \centering
    \resizebox{\textwidth}{!}{
    \begin{tabular}{l|c|cc|cc|c|c|cc|cc|c|c}
    \Xhline{1pt}
    \multirow{2}{*}{\textbf{Methods}} & \multirow{2}{*}{\textbf{Ref}} & \multicolumn{6}{c|}{\textbf{Light Degradation}} & \multicolumn{6}{c}{\textbf{Heavy Degradation}} \\
    \cmidrule(lr){3-14} 
    & & \textbf{PSNR} $\uparrow$ & \textbf{SSIM} $\uparrow$ & \textbf{LPIPS} $\downarrow$ & \textbf{MUSIQ} $\uparrow$ & \textbf{LMSE} $\downarrow$ & \textbf{ID} $\uparrow$ & \textbf{PSNR} $\uparrow$ & \textbf{SSIM} $\uparrow$ & \textbf{LPIPS} $\downarrow$ & \textbf{MUSIQ} $\uparrow$ & \textbf{LMSE} $\downarrow$ & \textbf{ID} $\uparrow$ \\    
    \Xhline{1pt}
    DR2 + SPAR \cite{DR22023} & \xmark & \color{blue}{25.21} & \red{0.750} & 0.193 & 44.85 & 2.831 & 0.711 & 21.52 & \red{0.703} & 0.289 & 20.53 & 6.479 & 0.385 \\
     CodeFormer \cite{CodeFormer2022} & \xmark & 25.03 & 0.714 & 0.136 & \color{blue}{75.38} & 2.498 & 0.774 & 21.16 & 0.641 & \color{blue}{0.196} & 73.55 & 4.793 & 0.379 \\
     ASFFNet \cite{ASFFNet2020} & \cmark & 25.07 & 0.742 & \red{0.127} & 72.09 & 2.483 & 0.843 & 21.25 & 0.641 & 0.199 & 71.55 & 11.661 & 0.399 \\
     DMDNet \cite{DMDnet2022} & \cmark & 23.97 & 0.715 & 0.158 & 70.31 & 2.637 & \color{blue}{0.867} & 21.39 & \color{blue}{0.652} & 0.211 & 67.85 & 6.966 & 0.450 \\
     PFStorer \cite{PFStorer2024} & \cmark & 25.16 & 0.685 & 0.136 & \red{76.12} & \color{blue}{2.230} & 0.853 & \red{22.31} & 0.638 & \red{0.182} & \red{75.44} & \red{3.918} & \color{blue}{0.473} \\
    \Xhline{1pt}
     \textbf{Ours} & \cmark & \red{26.03} & \color{blue}{0.744} & \color{blue}{0.132} & 71.62 & \red{2.210} & \red{0.867} & \color{blue}{21.87} & 0.621 & {0.204} & \color{blue}{74.79} & \color{blue}{4.348} & \red{0.548} \\
    \Xhline{1pt}
    \end{tabular}
    }
    \caption{Quantitative comparison on light and heavy degradation levels. \red{Red} indicates the best and {\color{blue}{blue}} indicates the second best.}
    \vspace{-1.5mm}
    \label{tab: quantitive-results}
\end{table*}

\subsection{Adaptive ID-Scale Adjusting} \label{sec:adaptive id-scale adjusting}
In \cref{eq:ipa}, ID-Scale $\lambda$ is the hyperparameter that adjusts the ID injection degree, balancing the generation freedom and ID preservation. Although we fix $\lambda=0.75$ in the training stage, we still find that varying degrees of ID prior injection influence the accuracy of the restored images during the inference stage. As shown in \cref{fig:IncreasingIPS}, under light degradation conditions, a high ID-Scale easily causes the recovered faces to be inaccurate. For example, as highlighted in red boxes, the generated images with ID-Scale=0.5 and 1.0 exhibit deep winkles that are not present in the ground-truth. Conversely, under heavy degradation conditions, a low ID-Scale is insufficient for ID preservation. Therefore, we design an Adaptive ID-Scale Adjusting strategy based on the degradation levels of the LQ images. The strategy should adhere to the following rule: ID-Scale should increase as the level of degradation rises.

We restore 150 LQ images with varying levels of degradation using our method, with ID-Scale ranging from 0 to 2 with interval of 0.04. We adopt the MUSIQ metric \cite{MUSIQ2021} to quantify the degradation level of images. A higher MUSIQ value indicates a smaller level of degradation. \cref{fig:IDScaleCurve} illustrates the curves of the average ID similarity (using ArcFace \cite{ArcFace2019}) values with respect to ID-Scale across different MUSIQ intervals. It can be observed that when MUSIQ is high, a smaller ID-Scale is beneficial for restoration; When MUSIQ is low, ID similarity initially rises and then decreases with increasing ID-Scale. We identify the optimal ID-Scale values for different MUSIQ levels and empirically use the following formula to adjust ID-Scale $\lambda$,
\begin{equation}
\lambda = \exp\left(\frac{\alpha-{\rm MUSIQ}(\mathbf{I}_{\text{lq}})}{\beta}\right),
 \label{eq:IPSA}
\end{equation}
where we set $\alpha=9.5$ and $\beta=10$.

%% file: sec/4_experiment.tex
\section{Experiments} \label{sec:experiment}
\subsection{Experimental Settings}

\noindent\textbf{Datasets.}
For the base model training, we use FFHQ \cite{FFHQ2019} and VGGFace2 \cite{VGGFace22018} datasets. For the FIR-Adapter training, we select 9,384 identities from the VGGFace2 and Celeb-Ref \cite{CelebRef2022} datasets, with each identity comprising 5-40 images. Additionally, we clean the training dataset by removing low-quality facial images using ArcFace backbone \cite{ArcFace2019}. For synthetic data evaluation, we select 50 identities from the remaining Celeb-Ref datasets, randomly choosing 2 images for each identity as the ground truth and reference images. We introduce two levels of degradation, \textit{i.e.}, light and heavy, to obtain LQ input images for comprehensive evaluation. For real-world data evaluation, we collect LQ and HQ images of 20 identities from the Internet.  

\noindent\textbf{Implement Details.} Our RestorerID is built based on the Stable Diffusion v1.5-base. We train the base model for 60,000 iterations and FIR-Adapter for 30,000 iterations with a batch size of 16. We use AdamW \cite{AdamW2017} optimizer and the learning rate is set to $5 \times 10^{-5}$. The training process is conducted on $512\times512$ resolution with 2 NVIDIA 48G-A6000 GPUs. For inference, we adopt DDIM \cite{DDIM2020} sampling with 50 timesteps and use classifier-free guidance (CFG) \cite{CFG2022} to guide the denoising process with $\lambda_{\text{cfg}}=7.5$.

\noindent\textbf{Evaluation Metrics.}
We use PSNR, SSIM, LPIPS \cite{LPIPS2018}, MUSIQ \cite{MUSIQ2021}, LMSE (Landmark MSE) \cite{LMSE2022}, and ID (cosine similarity with ArcFace\cite{ArcFace2019}) as evaluation metrics.

\noindent\textbf{Comparing Methods.} We compare our method with both reference-guided and blind face restoration approaches. The reference-guided approaches include ASFFNet \cite{ASFFNet2020}, DMDNet \cite{DMDnet2022}, and PFStorer \cite{PFStorer2024}, while the blind face restoration approaches include CodeFormer \cite{CodeFormer2022} and DR2 + SPAR \cite{DR22023}. Note that PFStorer is a test-tuning approach and it employs 5 reference images for each identity tuning.


\begin{figure*}[!t]
    \centering
    \begin{overpic}[width=\linewidth]{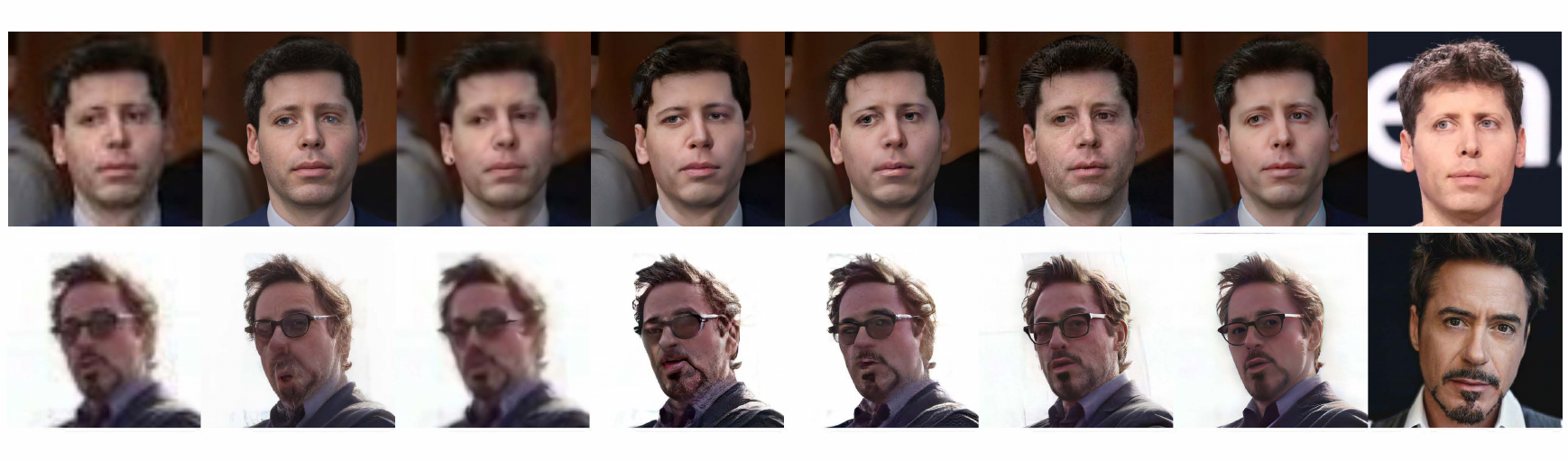}
        \put(6, 0){\footnotesize Input}
        \put(14, 0){\footnotesize DR2 + SPAR~\cite{DR22023}}
        \put(27, 0){\footnotesize CodeFormer~\cite{CodeFormer2022}}
        \put(40, 0){\footnotesize ASFFNet~\cite{ASFFNet2020}}
        \put(52, 0){\footnotesize DMDNet~\cite{DMDnet2022}}
        \put(65, 0){\footnotesize PFStorer~\cite{PFStorer2024}}
        \put(80, 0){\footnotesize \textbf{Ours}}
        \put(90, 0){\footnotesize Pseudo-GT}
    \end{overpic}
    \caption{Qualitative comparison with state-of-the-art restoration models on real-world images. Images are sourced from the Internet.}
    \vspace{-3mm}
    \label{fig: real-world}
\end{figure*}
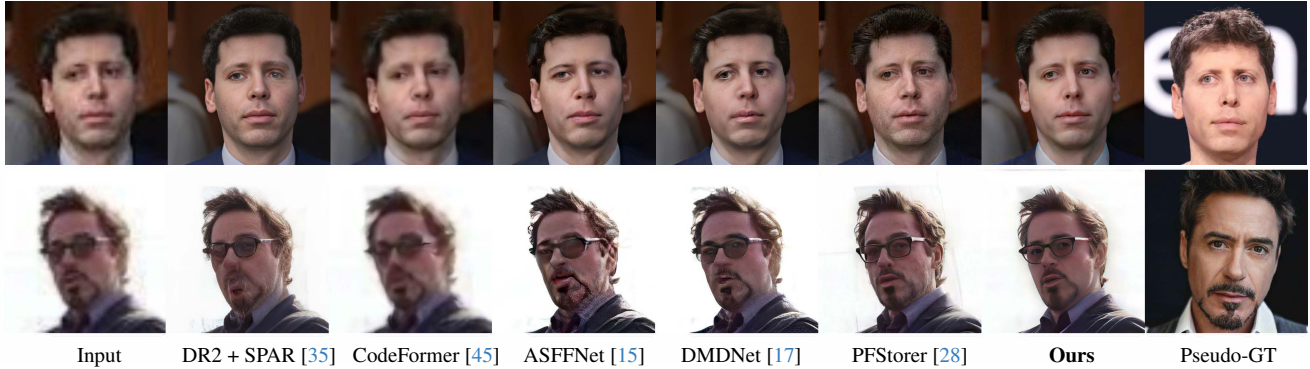

\subsection{Performance Comparison}
\noindent\textbf{Quantitative.}
\cref{tab: quantitive-results} lists the quantitative results of RestorerID and other competitors on light and heavy degradation levels.
From the results, it can be observed that RestorerID achieves competitive performance on PSNR, SSIM, and LPIPS metrics in the light degradation scenario, obtaining a PSNR of 26.03, which ranks among the top results. 
For the heavy degradation scenario, blind face restoration approaches demonstrate superior performance on SSIM. We hypothesize that this is because the SSIM metric emphasizes image sharpness and structural clarity. However, it overlooks the fidelity of facial details and naturalness, which are critical for a realistic face restoration.
Despite this, our method still produces visually pleasing results in terms of facial identity preservation and overall landmark consistency.
Notably, RestorerID achieves the best performance in ID metric for both light and heavy distortions, which underscores the ability of our model to retain personalized facial characteristics. In particular, under heavy degradation, our model surpasses the second-best competitor by 0.075. These results highlight the strength of RestorerID in accurately extracting personalized features from the reference image and leveraging these features for effective face restoration, even in challenging scenarios where input quality is severely degraded.

\begin{figure}
    \centering
    \includegraphics[width=1.0\linewidth]{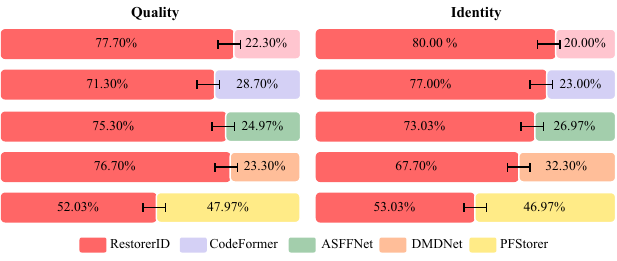}
    \caption{\textbf{User study results}. Voting rates of our RestorerID compared with other approaches on Quality and Identity metrics.}
    \label{fig: user-study}
    \vspace{-4mm}
\end{figure}

\noindent\textbf{Qualitative.} 
\cref{fig: CelebRef-Heavy} presents the visual results of face restoration at the heavy degradation level across different approaches. It is observed that our RestorerID outperforms other approaches, particularly in ID preservation. Specifically, when the noisy input images are difficult to recognize due to blurred facial features such as the eyes and nose, our method can restore the face and preserve the ID more effectively than blind face restoration approaches. Compared to PFStorer, our method also demonstrates superior identity preservation on features like the eyes and nose, as shown in the first row of \cref{fig: CelebRef-Heavy}.



\begin{figure*}[!h]
    \centering
    \begin{overpic}[width=0.98\linewidth]{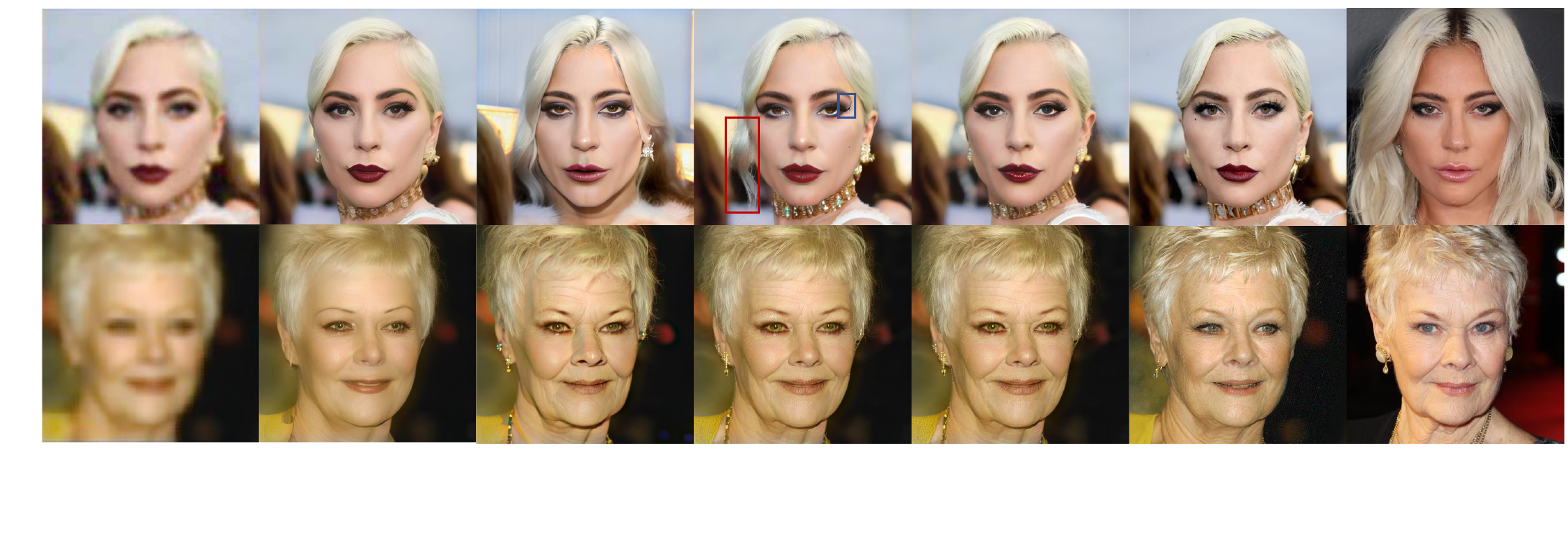}
        \put(7, 4){\footnotesize LQ Input}
        \put(20, 4){\footnotesize Base Model}
        \put(32, 4){\footnotesize + ID Injection}
        \put(46, 4){\footnotesize + ID Injection}
        \put(46, 2){\footnotesize + FIR-Adapter}
        \put(60, 4){\footnotesize + ID Injection}
        \put(60, 2){\footnotesize + FIR-Adapter}
        \put(60, 0){\footnotesize + AIDSA~(Ours)}
        \put(77, 4){\footnotesize GT}
        \put(92, 4){\footnotesize Ref}
        \put(, 7){\rotatebox{90}{\scriptsize Heavy Degradation}}
        \put(1, 21){\rotatebox{90}{\scriptsize Light Degradation}}
    \end{overpic}
    \caption{The qualitative results of ablation studies on the proposed components under light and heavy degradations. We incrementally add the ID injection, FIR-Adapter, and Adaptive ID-Scale Adjusting strategy (AIDSA) to the base model. Red and blue boxes highlight the haircut and eyebrow inconsistency.}
    \label{fig: Ablation1}
    \vspace{-3.5mm}
\end{figure*}

\begin{figure}[!h]
    \centering
    \begin{overpic}[width=0.98\linewidth]{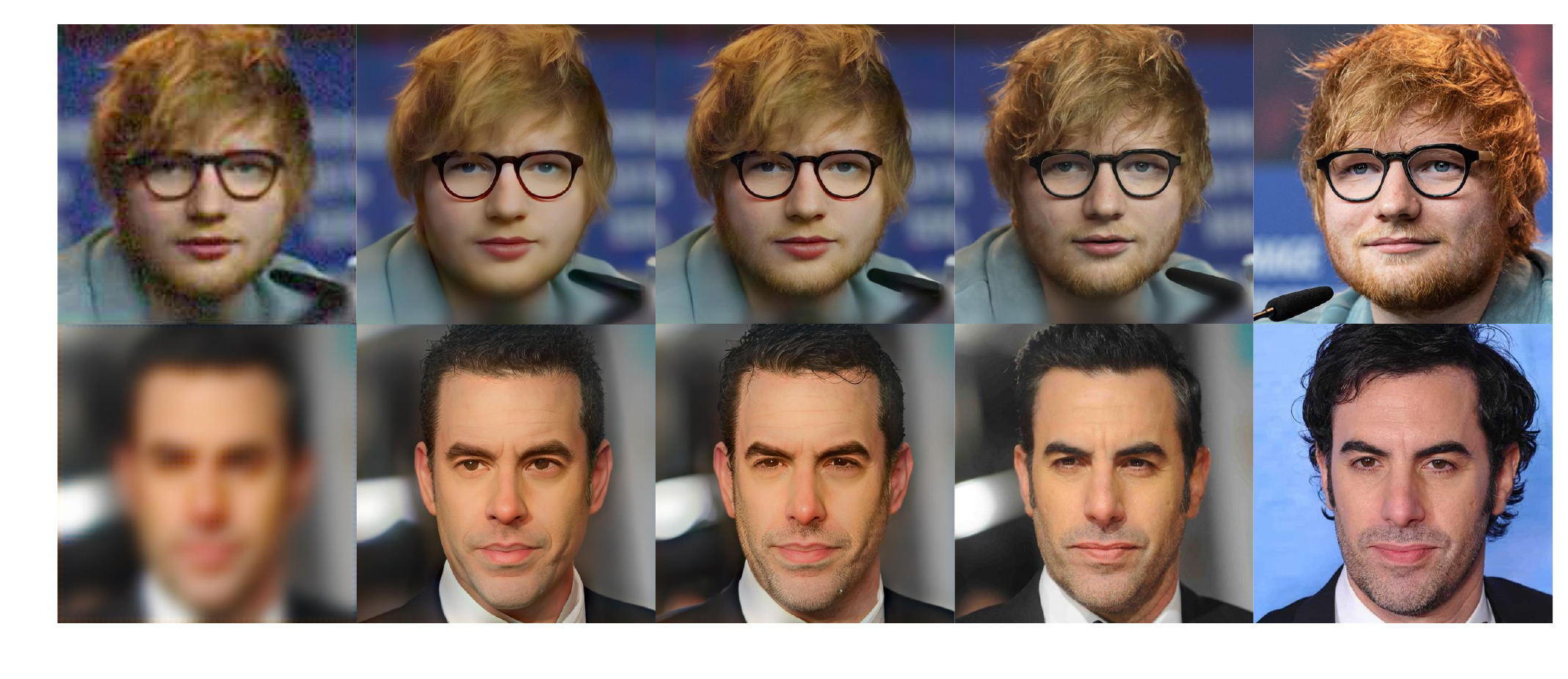}
    \put(9, 1){\tiny LQ Input}
    \put(23, 1){\tiny Single-Stage Training}
    \put(49, 1){\tiny Ours}
    \put(69, 1){\tiny GT}
    \put(88, 1){\tiny Ref}
    \put(1, 4){\rotatebox{90}{\tiny Heavy Degradation}}
    \put(1, 24){\rotatebox{90}{\tiny Light Degradation}}
    \end{overpic}
    \vspace{-2mm}
    \caption{The qualitative results of ablation study on the two-stage training strategy under light and heavy degradations.}
    \label{fig: Ablation2}
    \vspace{-3mm}
\end{figure}
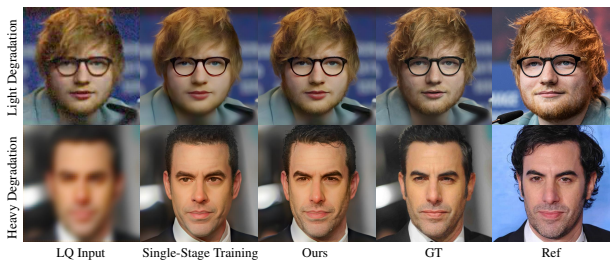

We also evaluate our RestorerID on real-world scenarios, as shown in \cref{fig: real-world}. 
Our method continues to demonstrate superior performance in both ID preservation and image quality.
In comparison, blind face restoration approaches such as DR2+SPAR and CodeFormer struggle to restore images faithfully, while reference-guided approaches face challenges in accurately recovering fine facial details. Additionally, PFStorer often produces artifacts and distortions in facial regions, highlighting the robustness of our approach. From the results, we can conclude that our RestorerID is also effective in real-world scenarios.



\begin{table}[!t]
    \centering
    \scalebox{0.75}{
    \begin{tabular}{c|ccc|ccc}
    \Xhline{1pt}
    \multicolumn{1}{c|}{Degradation} & \multicolumn{3}{c|}{light} & \multicolumn{3}{c}{heavy} \\
    \Xhline{1pt}
    \multicolumn{1}{c|}{\textbf{}} & LPIPS$\downarrow$ & LMSE$\downarrow$ & \multicolumn{1}{c|}{ID$\uparrow$} & {LPIPS$\downarrow$} &{LMSE$\downarrow$} & \multicolumn{1}{c}{ID$\uparrow$} \\  
    \Xhline{1pt}
     Base Model & {0.141}&{2.214}&{0.859}    &{0.206} &{4.392} &{0.401} \\
     + ID Injection &{0.158}&{2.571}&{0.831} &{0.221} &{5.635} &\textbf{0.557} \\
     + FIR Adapter &{0.141}&{2.315}&{0.859} &{0.207} &{4.427} &{0.552} \\
     + AIDSA (Ours) &\textbf{0.132}&\textbf{2.210}&\textbf{0.867} &\textbf{0.204} &\textbf{4.348} &{0.548} \\
     \Xhline{1pt}
    \end{tabular}
    }
    \caption{Ablation study on the proposed components, including ID injection, FIR-Adapter, and Adaptive ID-Scale Adjusting (AIDSA). The best ones are highlighted in bold.}
    \label{tab:Ablation1}
    \vspace{-3mm}
\end{table}

\noindent\textbf{User Study.} We conduct a user study to assess human preferences for perceptual quality, complementing the quantitative metrics.
Totally 30 participants evaluate 100 images on two aspects: quality and identity fidelity relative to a reference image. The collected data includes images with varying levels of degradation, comprising 25\% with light degradation and 75\% with heavy degradation. The results are shown in \cref{fig: user-study}. It can be observed that our method received the highest number of votes for both image quality and identity fidelity.
Specifically, RestorerID outperforms blind face restoration methods in both metrics and surpasses reference-guided methods such as ASFFNet and DMDNet in large margin, while achieving competitive results compared to the test-tuning method PFStorer. These findings demonstrate that RestorerID effectively delivers high-quality face restoration.

\subsection{Ablation Studies}
\noindent\textbf{The Effective of Proposed Components.} 
To assess the impact of each component on model performance, we incrementally add the ID Injection, FIR-Adapter, and Adaptive ID-Scale Adjusting strategy (AIDSA) components to the baseline model and conduct experiments under varying levels of degradation. \cref{tab:Ablation1} presents the LPIPS, LMSE, and ID metrics for each model setting.
The results show that incorporating the ID Injection component significantly improves the ID metric from 0.401 to 0.557 under heavy degradation, indicating a substantial enhancement in ID preservation. However, this also results in a decline in image quality.
Furthermore, the introduction of the FIR-Adapter aids in balancing ID preservation with image quality. This results in an ID metric increase from 0.831 to 0.859 under light degradation. It also leads to an LMSE metric reduction from 5.635 to 4.427 under heavy degradation. These improvements signify enhanced ID preservation while maintaining image quality.
Additionally, incorporating the AIDSA strategy yields the best results in both image quality and identity preservation for face image restoration.

\cref{fig: Ablation1} illustrates the qualitative results of the ablation study on the proposed components. It is observed that the adoption of ID injection component successfully incorporates the reference prior into the output, especially under heavy degradation, while it diminishes the structure consistency under light degradation. The adoption of FIR-Adapter effectively improves the image quality while maintaining ID similarity. But there still exists some content inconsistency, as highlighted in red and blue boxes. The addition of AIDSA further addresses the remaining issues, providing a more faithful restoration outcome. 

\noindent\textbf{The Necessity of Two-Stage Training.}
To verify the necessity of the two-stage training, we conducted a single-stage training by combining the base model and FIR-Adapter training together under the same settings. As shown in \cref{tab:Ablation2}, the performance of the single-stage training model is generally inferior to that of the two-stage training model. For example, under heavy degradation, the ID metric for single-stage training is 0.470, compared to 0.548 for two-stage training, and the LMSE is 4.626, higher than 4.348 for two-stage training. This is because the two-stage training for the base model and FIR-Adapter more effectively explores their capabilities of blind face restoration and information rebalancing, respectively, which is crucial for improving model performance.

\cref{fig: Ablation2} illustrates the qualitative results of the ablation study on two-stage training strategy. It is observed that the images produced by the single-stage training model show deficiencies in facial details, such as the beard and eyes. In contrast, our two-stage training model achieves superior performance in both image quality and ID preservation.

\begin{table}[!t]
    \centering
    \scalebox{0.75}{
    \begin{tabular}{c|ccc|ccc}
    \Xhline{1pt}
    \multicolumn{1}{c|}{Degradation} & \multicolumn{3}{c|}{light} & \multicolumn{3}{c}{heavy} \\
    \Xhline{1pt}
    \multicolumn{1}{c|}{\textbf{}} & LPIPS$\downarrow$ & LMSE$\downarrow$ & \multicolumn{1}{c|}{ID$\uparrow$} & {LPIPS$\downarrow$} &{LMSE$\downarrow$} & \multicolumn{1}{c}{ID$\uparrow$} \\  
    \Xhline{1pt}
     Single-Stage &{0.137}&{2.225}&{0.852}  &\textbf{0.195} &{4.626} &{0.470} \\
     Ours &\textbf{0.132}&\textbf{2.210}&\textbf{0.867} &{0.204} &\textbf{4.348} &\textbf{0.548} \\
     \Xhline{1pt}
    \end{tabular}
    }
    \caption{Ablation study for our two-stage training strategy.}
    \label{tab:Ablation2}
    \vspace{-3mm}
\end{table}




%% file: sec/5_conclusion.tex
\section{Conclusions} \label{sec:conclusion}
In this work, we have introduced RestorerID, a tuning-free face restoration method with ID preservation, capable of recovering high-quality face images across varying levels of degradation. RestorerID combines the face restoration and ID preservation into a unified stable diffusion-based framework by adopting the independent LQ spatial model and ID model. To balance the LQ structural and reference ID information, we propose the FIR-Adapter and Adaptive ID-Scale Adjusting strategy, which seamlessly enhance the restoration quality with ID-preserving capabilities. Experimental results on Celeb-Ref and real-world scenario datasets proves that RestorerID achieves excellent face restoration with ID preservation, even in challenging heavy-degradation scenarios. We hope this work can contribute to advancements in high-fidelity photography.

%% file: sec/6_appendix.tex

%

%% file: main.bbl
\begin{thebibliography}{45}
\providecommand{\natexlab}[1]{#1}
\providecommand{\url}[1]{\texttt{#1}}
\expandafter\ifx\csname urlstyle\endcsname\relax
  \providecommand{\doi}[1]{doi: #1}\else
  \providecommand{\doi}{doi: \begingroup \urlstyle{rm}\Url}\fi

\bibitem[Betker et~al.(2023)Betker, Goh, Jing, Brooks, Wang, Li, Ouyang,
  Zhuang, Lee, Guo, et~al.]{dalle3}
James Betker, Gabriel Goh, Li Jing, Tim Brooks, Jianfeng Wang, Linjie Li, Long
  Ouyang, Juntang Zhuang, Joyce Lee, Yufei Guo, et~al.
\newblock Improving image generation with better captions.
\newblock \emph{Computer Science. https://cdn. openai. com/papers/dall-e-3.
  pdf}, 2:\penalty0 3, 2023.

\bibitem[Cao et~al.(2018)Cao, Shen, Xie, Parkhi, and Zisserman]{VGGFace22018}
Qiong Cao, Li Shen, Weidi Xie, Omkar~M Parkhi, and Andrew Zisserman.
\newblock Vggface2: A dataset for recognising faces across pose and age.
\newblock In \emph{2018 13th IEEE international conference on automatic face \&
  gesture recognition (FG 2018)}, pages 67--74. IEEE, 2018.

\bibitem[Chen et~al.(2023)Chen, Huang, Liu, Shen, Zhao, and Zhao]{anydoor}
Xi Chen, Lianghua Huang, Yu Liu, Yujun Shen, Deli Zhao, and Hengshuang Zhao.
\newblock Anydoor: Zero-shot object-level image customization.
\newblock \emph{CoRR}, abs/2307.09481, 2023.

\bibitem[Chen et~al.(2024)Chen, Tan, Wang, Zhang, Luo, and Cao]{BFRffusion2024}
Xiaoxu Chen, Jingfan Tan, Tao Wang, Kaihao Zhang, Wenhan Luo, and Xiaochun Cao.
\newblock Towards real-world blind face restoration with generative diffusion
  prior.
\newblock \emph{IEEE Transactions on Circuits and Systems for Video
  Technology}, 2024.

\bibitem[Deng et~al.(2019)Deng, Guo, Xue, and Zafeiriou]{ArcFace2019}
Jiankang Deng, Jia Guo, Niannan Xue, and Stefanos Zafeiriou.
\newblock Arcface: Additive angular margin loss for deep face recognition.
\newblock In \emph{CVPR}, pages 4690--4699, 2019.

\bibitem[Gal et~al.(2023)Gal, Alaluf, Atzmon, Patashnik, Bermano, Chechik, and
  Cohen{-}Or]{textual_inversion}
Rinon Gal, Yuval Alaluf, Yuval Atzmon, Or Patashnik, Amit~Haim Bermano, Gal
  Chechik, and Daniel Cohen{-}Or.
\newblock An image is worth one word: Personalizing text-to-image generation
  using textual inversion.
\newblock In \emph{ICLR}, 2023.

\bibitem[Guo et~al.(2019)Guo, Yan, Zhang, Zuo, and Zhang]{CBDNet2019}
Shi Guo, Zifei Yan, Kai Zhang, Wangmeng Zuo, and Lei Zhang.
\newblock Toward convolutional blind denoising of real photographs.
\newblock In \emph{CVPR}, pages 1712--1722, 2019.

\bibitem[Ho and Salimans(2022)]{CFG2022}
Jonathan Ho and Tim Salimans.
\newblock Classifier-free diffusion guidance.
\newblock \emph{arXiv preprint arXiv:2207.12598}, 2022.

\bibitem[Ho et~al.(2020)Ho, Jain, and Abbeel]{DDPM2020}
Jonathan Ho, Ajay Jain, and Pieter Abbeel.
\newblock Denoising diffusion probabilistic models.
\newblock In \emph{NeurIPS}, pages 6840--6851, 2020.

\bibitem[Hu et~al.(2021)Hu, Shen, Wallis, Allen-Zhu, Li, Wang, Wang, and
  Chen]{lora}
Edward~J Hu, Yelong Shen, Phillip Wallis, Zeyuan Allen-Zhu, Yuanzhi Li, Shean
  Wang, Lu Wang, and Weizhu Chen.
\newblock Lora: Low-rank adaptation of large language models.
\newblock \emph{arXiv preprint arXiv:2106.09685}, 2021.

\bibitem[Karras et~al.(2019)Karras, Laine, and Aila]{FFHQ2019}
Tero Karras, Samuli Laine, and Timo Aila.
\newblock A style-based generator architecture for generative adversarial
  networks.
\newblock In \emph{CVPR}, pages 4401--4410, 2019.

\bibitem[Ke et~al.(2021)Ke, Wang, Wang, Milanfar, and Yang]{MUSIQ2021}
Junjie Ke, Qifei Wang, Yilin Wang, Peyman Milanfar, and Feng Yang.
\newblock Musiq: Multi-scale image quality transformer.
\newblock In \emph{ICCV}, pages 5148--5157, 2021.

\bibitem[Kumari et~al.(2023)Kumari, Zhang, Zhang, Shechtman, and
  Zhu]{customdiffusion}
Nupur Kumari, Bingliang Zhang, Richard Zhang, Eli Shechtman, and Jun{-}Yan Zhu.
\newblock Multi-concept customization of text-to-image diffusion.
\newblock In \emph{CVPR}, pages 1931--1941. {IEEE}, 2023.

\bibitem[Li et~al.(2018)Li, Liu, Ye, Zuo, Lin, and Yang]{GFRNet2018}
Xiaoming Li, Ming Liu, Yuting Ye, Wangmeng Zuo, Liang Lin, and Ruigang Yang.
\newblock Learning warped guidance for blind face restoration.
\newblock In \emph{ECCV}, pages 272--289, 2018.

\bibitem[Li et~al.(2020)Li, Li, Ren, Zhang, Wang, and Zuo]{ASFFNet2020}
Xiaoming Li, Wenyu Li, Dongwei Ren, Hongzhi Zhang, Meng Wang, and Wangmeng Zuo.
\newblock Enhanced blind face restoration with multi-exemplar images and
  adaptive spatial feature fusion.
\newblock In \emph{CVPR}, pages 2706--2715, 2020.

\bibitem[Li et~al.(2022{\natexlab{a}})Li, Zhang, Zhou, Zhang, and
  Zuo]{CelebRef2022}
Xiaoming Li, Shiguang Zhang, Shangchen Zhou, Lei Zhang, and Wangmeng Zuo.
\newblock Learning dual memory dictionaries for blind face restoration.
\newblock \emph{IEEE Transactions on Pattern Analysis and Machine
  Intelligence}, 45\penalty0 (5):\penalty0 5904--5917, 2022{\natexlab{a}}.

\bibitem[Li et~al.(2022{\natexlab{b}})Li, Zhang, Zhou, Zhang, and
  Zuo]{DMDnet2022}
Xiaoming Li, Shiguang Zhang, Shangchen Zhou, Lei Zhang, and Wangmeng Zuo.
\newblock Learning dual memory dictionaries for blind face restoration.
\newblock \emph{T-PAMI}, 45\penalty0 (5):\penalty0 5904--5917,
  2022{\natexlab{b}}.

\bibitem[Li et~al.(2024)Li, Cao, Wang, Qi, Cheng, and Shan]{photomaker}
Zhen Li, Mingdeng Cao, Xintao Wang, Zhongang Qi, Ming-Ming Cheng, and Ying
  Shan.
\newblock Photomaker: Customizing realistic human photos via stacked id
  embedding.
\newblock In \emph{CVPR}, pages 8640--8650, 2024.

\bibitem[Loshchilov(2017)]{AdamW2017}
I Loshchilov.
\newblock Decoupled weight decay regularization.
\newblock \emph{arXiv preprint arXiv:1711.05101}, 2017.

\bibitem[Radford et~al.(2021)Radford, Kim, Hallacy, Ramesh, Goh, Agarwal,
  Sastry, Askell, Mishkin, Clark, et~al.]{CLIP2021}
Alec Radford, Jong~Wook Kim, Chris Hallacy, Aditya Ramesh, Gabriel Goh,
  Sandhini Agarwal, Girish Sastry, Amanda Askell, Pamela Mishkin, Jack Clark,
  et~al.
\newblock Learning transferable visual models from natural language
  supervision.
\newblock In \emph{ICML}, pages 8748--8763, 2021.

\bibitem[Rombach et~al.(2022{\natexlab{a}})Rombach, Blattmann, Lorenz, Esser,
  and Ommer]{SD2022}
Robin Rombach, Andreas Blattmann, Dominik Lorenz, Patrick Esser, and Bj{\"o}rn
  Ommer.
\newblock High-resolution image synthesis with latent diffusion models.
\newblock In \emph{CVPR}, pages 10684--10695, 2022{\natexlab{a}}.

\bibitem[Rombach et~al.(2022{\natexlab{b}})Rombach, Blattmann, Lorenz, Esser,
  and Ommer]{stable_diffusion}
Robin Rombach, Andreas Blattmann, Dominik Lorenz, Patrick Esser, and
  Bj{\"{o}}rn Ommer.
\newblock High-resolution image synthesis with latent diffusion models.
\newblock In \emph{CVPR}, pages 10674--10685, 2022{\natexlab{b}}.

\bibitem[Ronneberger et~al.(2015)Ronneberger, Fischer, and Brox]{UNet2015}
Olaf Ronneberger, Philipp Fischer, and Thomas Brox.
\newblock U-net: Convolutional networks for biomedical image segmentation.
\newblock In \emph{MICCAI}, pages 234--241, 2015.

\bibitem[Ruiz et~al.(2023)Ruiz, Li, Jampani, Pritch, Rubinstein, and
  Aberman]{dreambooth}
Nataniel Ruiz, Yuanzhen Li, Varun Jampani, Yael Pritch, Michael Rubinstein, and
  Kfir Aberman.
\newblock Dreambooth: Fine tuning text-to-image diffusion models for
  subject-driven generation.
\newblock In \emph{CVPR}, pages 22500--22510, 2023.

\bibitem[Song et~al.(2020)Song, Meng, and Ermon]{DDIM2020}
Jiaming Song, Chenlin Meng, and Stefano Ermon.
\newblock Denoising diffusion implicit models.
\newblock \emph{arXiv preprint arXiv:2010.02502}, 2020.

\bibitem[Valevski et~al.(2023)Valevski, Wasserman, Matias, and
  Leviathan]{valevski2023face0}
Dani Valevski, Danny Wasserman, Yossi Matias, and Yaniv Leviathan.
\newblock Face0: Instantaneously conditioning a text-to-image model on a face,
  2023.

\bibitem[Van Den~Oord et~al.(2017)Van Den~Oord, Vinyals, et~al.]{VAE2017}
Aaron Van Den~Oord, Oriol Vinyals, et~al.
\newblock Neural discrete representation learning.
\newblock In \emph{NeurIPS}, 2017.

\bibitem[Varanka et~al.(2024)Varanka, Toivonen, Tripathy, Zhao, and
  Acar]{PFStorer2024}
Tuomas Varanka, Tapani Toivonen, Soumya Tripathy, Guoying Zhao, and Erman Acar.
\newblock Pfstorer: Personalized face restoration and super-resolution.
\newblock In \emph{CVPR}, pages 2372--2381, 2024.

\bibitem[Wan et~al.(2020)Wan, Zhang, Chen, Zhang, Chen, Liao, and
  Wen]{OldPhoto2020}
Ziyu Wan, Bo Zhang, Dongdong Chen, Pan Zhang, Dong Chen, Jing Liao, and Fang
  Wen.
\newblock Bringing old photos back to life.
\newblock In \emph{CVPR}, pages 2747--2757, 2020.

\bibitem[Wang et~al.(2024{\natexlab{a}})Wang, Yue, Zhou, Chan, and
  Loy]{StableSR2024}
Jianyi Wang, Zongsheng Yue, Shangchen Zhou, Kelvin~CK Chan, and Chen~Change
  Loy.
\newblock Exploiting diffusion prior for real-world image super-resolution.
\newblock \emph{IJCV}, pages 1--21, 2024{\natexlab{a}}.

\bibitem[Wang et~al.(2024{\natexlab{b}})Wang, Bai, Wang, Qin, and
  Chen]{instanceid}
Qixun Wang, Xu Bai, Haofan Wang, Zekui Qin, and Anthony Chen.
\newblock Instantid: Zero-shot identity-preserving generation in seconds.
\newblock In \emph{ECCV}, 2024{\natexlab{b}}.

\bibitem[Wang et~al.(2024{\natexlab{c}})Wang, Bai, Wang, Qin, and
  Chen]{instantid}
Qixun Wang, Xu Bai, Haofan Wang, Zekui Qin, and Anthony Chen.
\newblock Instantid: Zero-shot identity-preserving generation in seconds.
\newblock \emph{arXiv preprint arXiv:2401.07519}, 2024{\natexlab{c}}.

\bibitem[Wang et~al.(2021{\natexlab{a}})Wang, Li, Zhang, and Shan]{GFPGAN2021}
Xintao Wang, Yu Li, Honglun Zhang, and Ying Shan.
\newblock Towards real-world blind face restoration with generative facial
  prior.
\newblock In \emph{CVPR}, pages 9168--9178, 2021{\natexlab{a}}.

\bibitem[Wang et~al.(2021{\natexlab{b}})Wang, Xie, Dong, and
  Shan]{RealESRGAN2021}
Xintao Wang, Liangbin Xie, Chao Dong, and Ying Shan.
\newblock Real-esrgan: Training real-world blind super-resolution with pure
  synthetic data.
\newblock In \emph{ICCV}, pages 1905--1914, 2021{\natexlab{b}}.

\bibitem[Wang et~al.(2023)Wang, Zhang, Zhang, Zheng, Zhou, Zhang, and
  Wang]{DR22023}
Zhixin Wang, Ziying Zhang, Xiaoyun Zhang, Huangjie Zheng, Mingyuan Zhou, Ya
  Zhang, and Yanfeng Wang.
\newblock Dr2: Diffusion-based robust degradation remover for blind face
  restoration.
\newblock In \emph{CVPR}, pages 1704--1713, 2023.

\bibitem[Wei et~al.(2023)Wei, Zhang, Ji, Bai, Zhang, and Zuo]{ELITE}
Yuxiang Wei, Yabo Zhang, Zhilong Ji, Jinfeng Bai, Lei Zhang, and Wangmeng Zuo.
\newblock {ELITE:} encoding visual concepts into textual embeddings for
  customized text-to-image generation.
\newblock In \emph{ICCV}, pages 15897--15907. {IEEE}, 2023.

\bibitem[Yang et~al.(2023)Yang, Zhou, Tao, and Loy]{PGDiff2023}
Peiqing Yang, Shangchen Zhou, Qingyi Tao, and Chen~Change Loy.
\newblock {PGD}iff: Guiding diffusion models for versatile face restoration via
  partial guidance.
\newblock 36, 2023.

\bibitem[Yang et~al.(2021)Yang, Ren, Xie, and Zhang]{GPEN2021}
Tao Yang, Peiran Ren, Xuansong Xie, and Lei Zhang.
\newblock Gan prior embedded network for blind face restoration in the wild.
\newblock In \emph{CVPR}, pages 672--681, 2021.

\bibitem[Ye et~al.(2023)Ye, Zhang, Liu, Han, and Yang]{ip-adapter}
Hu Ye, Jun Zhang, Sibo Liu, Xiao Han, and Wei Yang.
\newblock Ip-adapter: Text compatible image prompt adapter for text-to-image
  diffusion models.
\newblock \emph{CoRR}, abs/2308.06721, 2023.

\bibitem[Yu et~al.(2024)Yu, Gu, Li, Hu, Kong, Wang, He, Qiao, and
  Dong]{SUPIR2024}
Fanghua Yu, Jinjin Gu, Zheyuan Li, Jinfan Hu, Xiangtao Kong, Xintao Wang,
  Jingwen He, Yu Qiao, and Chao Dong.
\newblock Scaling up to excellence: Practicing model scaling for
  photo-realistic image restoration in the wild.
\newblock In \emph{CVPR}, pages 25669--25680, 2024.

\bibitem[Yue and Loy(2024)]{DifFace2024}
Zongsheng Yue and Chen~Change Loy.
\newblock Dif{F}ace: Blind face restoration with diffused error contraction.
\newblock \emph{IEEE Transactions on Pattern Analysis and Machine
  Intelligence}, pages 1--15, 2024.

\bibitem[Zhang et~al.(2018)Zhang, Isola, Efros, Shechtman, and Wang]{LPIPS2018}
Richard Zhang, Phillip Isola, Alexei~A Efros, Eli Shechtman, and Oliver Wang.
\newblock The unreasonable effectiveness of deep features as a perceptual
  metric.
\newblock In \emph{CVPR}, pages 586--595, 2018.

\bibitem[Zhang et~al.(2024)Zhang, Liu, Song, Wang, Tang, Yu, Li, Tang, Hu, Pan,
  and Jing]{ssr_encoder}
Yuxuan Zhang, Jiaming Liu, Yiren Song, Rui Wang, Hao Tang, Jinpeng Yu, Huaxia
  Li, Xu Tang, Yao Hu, Han Pan, and Zhongliang Jing.
\newblock Ssr-encoder: Encoding selective subject representation for
  subject-driven generation.
\newblock In \emph{CVPR}, pages 8069--8078, 2024.

\bibitem[Zheng et~al.(2022)Zheng, Yang, Zhang, Bao, Chen, Huang, Yuan, Chen,
  Zeng, and Wen]{LMSE2022}
Yinglin Zheng, Hao Yang, Ting Zhang, Jianmin Bao, Dongdong Chen, Yangyu Huang,
  Lu Yuan, Dong Chen, Ming Zeng, and Fang Wen.
\newblock General facial representation learning in a visual-linguistic manner.
\newblock In \emph{CVPR}, pages 18697--18709, 2022.

\bibitem[Zhou et~al.(2022)Zhou, Chan, Li, and Loy]{CodeFormer2022}
Shangchen Zhou, Kelvin Chan, Chongyi Li, and Chen~Change Loy.
\newblock Towards robust blind face restoration with codebook lookup
  transformer.
\newblock In \emph{NeurIPS}, pages 30599--30611, 2022.

\end{thebibliography}
